%% file: bare_conf.tex
\documentclass[10pt, conference, compsocconf]{IEEEtran}
\ifCLASSINFOpdf
  \usepackage[pdftex]{graphicx}
\else
\fi
\usepackage[cmex10]{amsmath}
\usepackage{amssymb}
\input{defs}

\usepackage[implicit=false, bookmarks=false, hidelinks]{hyperref}

\begin{document}
%
\title{Linking Art through Human Poses}


\author{\IEEEauthorblockN{Tomas Jenicek \href{https://orcid.org/0000-0002-1372-6746}{\includegraphics[scale=0.25]{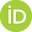}}, Ond{\v r}ej Chum \href{https://orcid.org/0000-0001-7042-1810}{\includegraphics[scale=0.25]{illustrations/orcid_32.png}}}
\IEEEauthorblockA{Visual Recognition Group, FEE, Czech Technical University in Prague\\
}
}


%


\maketitle

\input{illum.tex}

\section*{Acknowledgments}
This work was supported by the GA\v{C}R grant 19-23165S and the CTU student grant SGS17/185/OHK3/3T/13.



\input{biblio.bbl}

%
%
%

\end{document}

%% file: defs.tex
\usepackage{xspace}
\usepackage{times, float, amsfonts}
\usepackage{color, watermark}
\usepackage{tikz}

\newcommand{\ie}{{\em i.e.}\xspace}
\newcommand{\eg}{{\em e.g.}\xspace}

\newcommand{\T}{{\!\top}}

\def\oversim#1#2{\lower .5pt\vbox{\lineskiplimit=\maxdimen \lineskip=.5pt
    \ialign{$\mathsurround=0pt #1\hfil ##\hfil $\crcr #2\crcr \sim\crcr }}}

\newfloat{algorithm}{tbh}{loa}
\floatname{algorithm}{Algorithm}
\let \oldalgortihm = \algorithm
\def \algorithm {\oldalgortihm \noindent \hbox{} \hrulefill
\\[-12pt]
\let \oldcaption = \caption
\def \caption {\mbox{}\\[-15pt] \noindent \hbox{} \hrulefill \oldcaption}}

\definecolor{faded}{rgb}{.97,.97,.97}

%% file: illum.tex
\begin{abstract}
We address the discovery of composition transfer in artworks based on their visual content. Automated analysis of large art collections, which are growing as a result of art digitization among museums and galleries, is an important tool for art history and assists cultural heritage preservation. Modern image retrieval systems offer good performance on visually similar artworks, but fail in the cases of more abstract composition transfer. 
The proposed approach links artworks through a pose similarity of human figures depicted in images. 
Human figures are the subject of a large fraction of visual art from middle ages to modernity and their distinctive poses were often a source of inspiration among artists.
The method consists of two steps -- fast pose matching and robust spatial verification. We experimentally show that explicit human pose matching is superior to standard content-based image retrieval methods on a manually annotated art composition transfer dataset.

\end{abstract}

\begin{IEEEkeywords}
pose matching; art retrieval; inspiration discovery;
\end{IEEEkeywords}

\IEEEpeerreviewmaketitle

\section{Introduction}

In recent years, numerous efforts have been invested in digitization of art. This includes scanning of existing paper catalogues, such as in the Replica project~\cite{seguin2018new}, or acquiring new digital scans and photographs of artworks~\cite{guidi2015massive}. The digital collections will allow preservation and remote access to the cultural heritage, as well as efficient analysis of individual works or relations between them. In this paper, we focus on discovery of links between artworks based on visual appearance of figure arrangements in them.

\begin{figure}
    \centering
    \includegraphics[width=1.05in]{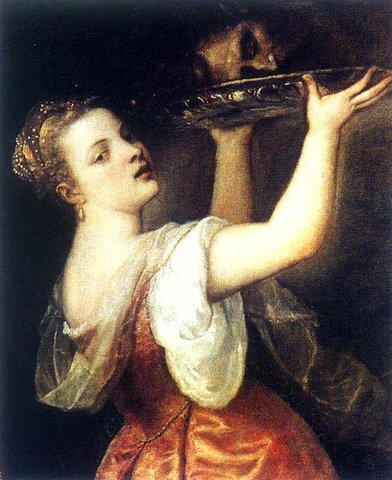}
    \includegraphics[width=1.05in]{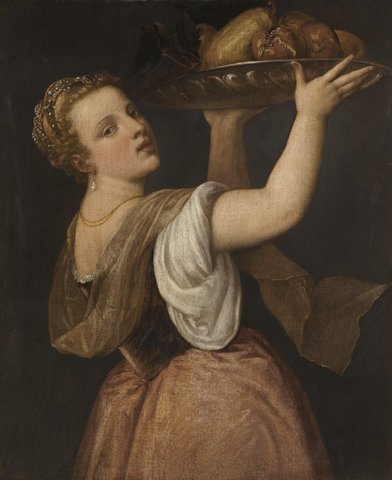}
    \includegraphics[width=1.05in]{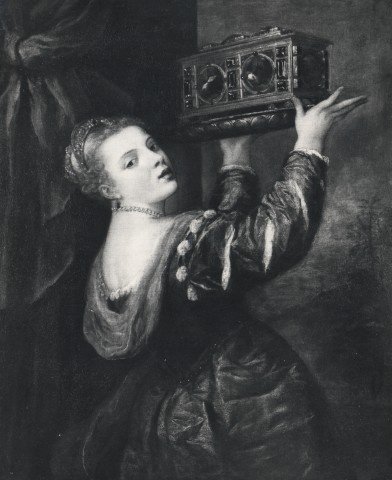}
    
    \vspace{0.1in}
    \includegraphics[width=1.6in]{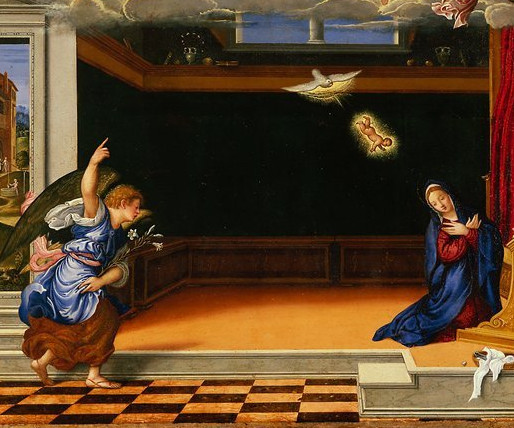}
    \includegraphics[width=1.6in]{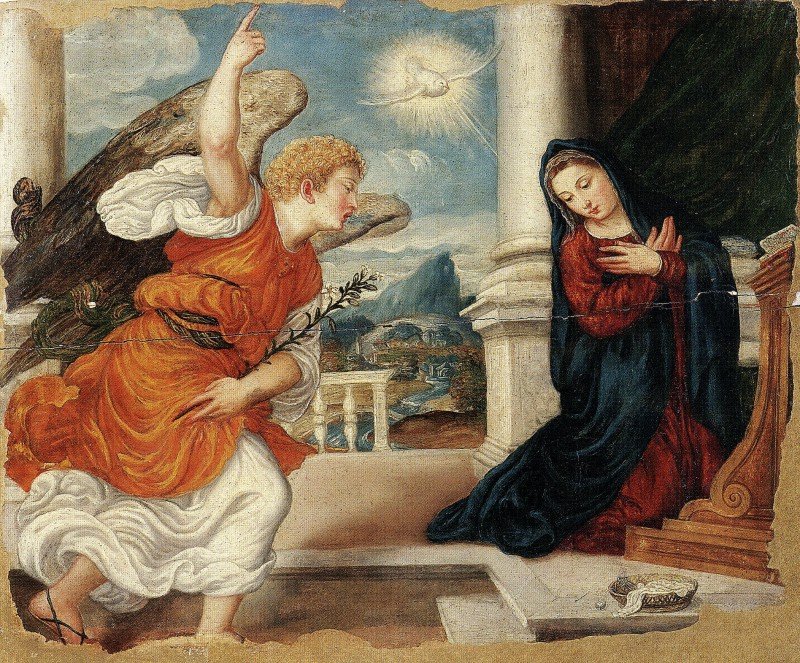}
    \caption{An example of human figure reuse in paintings. In the first row, a women with an identical posture and facial expression is painted in different contexts. In the first image, the head on the plate is of John the Baptists' and the painting belongs to Christian art. In other images, the woman is kept the same but the theme shifts away from religious art. In the second row, the theme is the same for both paintings as well as the posture of the two characters, but their mutual configuration changes together with their clothing and background.}
    \label{fig:motivation}
\end{figure}

Throughout the history, it was a common practice that artists took inspiration from each other to a varying degree.
Before Renaissance, the main value of art was in the materials used and authors were only craftsmen, so the variation between different artworks was low. The main purpose of art was depiction of the biblical themes and thus the same paintings were copied many times without any significant change. During Renaissance, this trend was gradually changed towards more abstract inspiration among painters. The workmanship itself started to be more valuable than just the materials, so the individuality of artists was increasing. One consequence of this was that some painters started copying a theme, such as the mutual configuration of characters, animals and objects, or the posture of individual characters, rather than the exact appearance which can then vary widely (see Figure~\ref{fig:motivation}).

\begin{figure}
    \centering
    \includegraphics[width=1.6in]{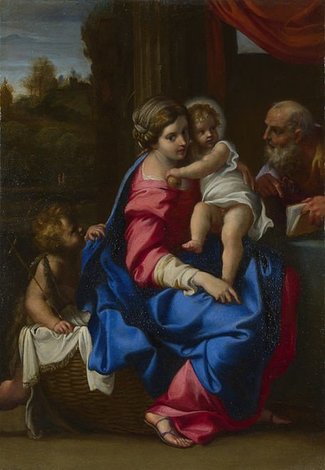}
    \includegraphics[width=1.6in]{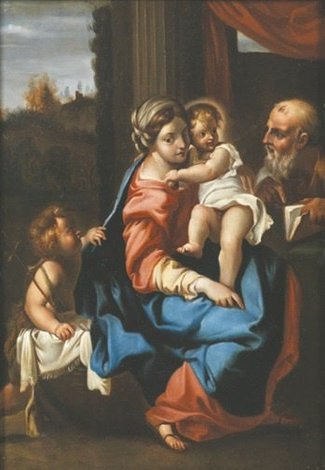}

    \vspace{0.05in}
    \includegraphics[width=1.6in]{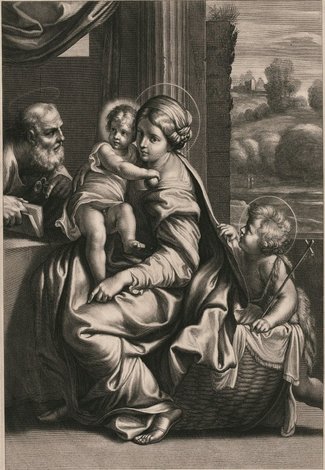}
    \includegraphics[width=1.6in]{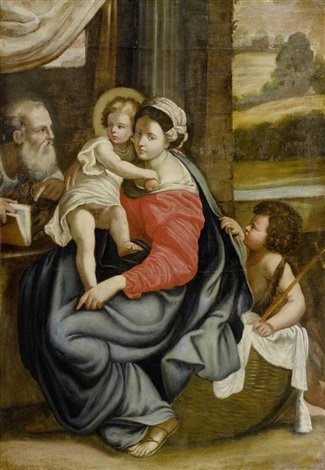}
    \caption{An example of discovered relations linking artworks in different stages of the copying process. The first two images represent a direct copy from an original. A careful comparison of the two images is necessary to confirm it is a replication created independently, not the same artwork in different stages of a restoration process. The third image is a print of an engraving. The fourth copy was most probably created from the print, so it is horizontally mirrored and the colors slightly differ, which is most notable on the curtain and on the bottom of Virgin Mary's robe. In this case, the authors tried to perform as precise replication as possible, thus minimizing variations between the paintings.}
    \label{fig:copy_pipeline}
\end{figure}

In planar visual art (paintings, drawings, prints and frescoes), we differentiate three types of similarity  -- physical link, replication and composition transfer. A physical link occurs between different machine copies of the same artwork. This appears in museum and gallery digitized collections where photographs of the same painting with a varying detail, or during different phases of restoration works, exist.
Before the invention of photography, the only means of making copies was by manually replicating the artworks. Paintings were often replicated by the author itself or by other artists in the same workshop, either by re-painting them or taking a drawing or an engraving.
An engraving was created for more popular paintings, and then used to print the outline of the original painting multiple times. The paintings consequently created from the print can be easily identified because they are horizontally flipped and their colors are different when compared to the original (an example can be seen in Figure~\ref{fig:copy_pipeline}).
Despite the efforts, there are visual differences between the original and the replica, even if the same medium was used. Finally, in the case of composition transfer, a painter incorporated a stylistic element after seeing the work of another artist. Elements being often copied include humans or groups of humans, but their layout can differ. This yielded a reuse of person poses and global compositions throughout the history.

This paper focuses on the discovery of an composition transfer among artworks containing humans using an image retrieval system. For each image in the database, the most similar images are found, creating a graph of artworks. The most valuable are the painting-drawing and painting-engraving relations, as the current image retrieval methods fail to connect such pairs because of the lack of visual similarity between different forms of the same artwork. More and more galleries and museums choose to digitize their collections and uncovering these relations allows interconnecting art databases without human intervention. Tracking the theme inspiration spread throughout time and space is invaluable for the art history field as it can provide a survey of popularity of a theme and its evolution as it spread.

\section{Prior Work}

Modern image retrieval methods, both local feature and CNN based, have been applied to search in datasets of artworks~\cite{seguin2016visual,seguin2017tracking}. This approach can be further improved with human-in-the-loop, as demonstrated by~\cite{seguin2018replica} where the annotations are iteratively refined and the CNN model is retrained. These systems have an ideal performance on physical links, and satisfactory performance on artworks with similar appearance and painting techniques. However, in the case of replication on different media and composition transfer through character poses, the image retrieval methods perform poorly.

Using visual recognition techniques on artworks is itself challenging. Various classification tasks are addressed in the literature, for example art theme classification~\cite{carneiro2012artistic}, artistic style classification~\cite{bar2014classification}, artist, genre, and artistic style joint classification~\cite{saleh2015large}, and artist, year, genre, artistic style and media joint classification~\cite{belhi2018towards}. Detection of people in artworks is implemented in~\cite{westlake2016detecting} and weakly supervised detection of objects in paintings is investigated in~\cite{gonthier2018weakly}. Multi-modal retrieval accross visual and textual representations of paintings is discussed by~\cite{garcia2018read}. The demand for art-related visual recognition algorithms is best demonstrated by the challenge of~\cite{mensink2014rijksmuseum}. This challenge consists of four different classification tasks on the digitized collection of Rijksmuseum in Amsterdam.

As the amount of annotated artworks is limited, some include also annotated datasets of natural images. Classification of both artistic and photography style on both photographs and paintings is addressed by~\cite{karayev2013recognizing}. In~\cite{crowley2014state, crowley2014search}, an object category classifier trained on natural images is generalized for artworks and spatial validation is defined for object retrieval. Different object detectors trained on natural images are evaluated on art in~\cite{redmon2016you}.

Another branch of research is focused on explicitly handling the domain gap between different visual domains, such as paintings, hand-drawn sketches and natural images. In~\cite{cai2015cross}, a number of different approaches to object classification across domains is tested, namely domain adaptation, part-based models and a CNN approach. An image similarity invariant to the visual domain is learned by~\cite{shrivastava2011data}. Another approach is to transform all images into a domain less sensitive to the visual domain. As shown by~\cite{Radenovic-ECCV18}, extracting edges and training a CNN only on edgemaps gives satisfactory results in image retrieval across visual domains.

\subsection{Pose retrieval}

One way to tackle the composition transfer discovery, also used in this paper, is through pose retrieval. There has been an ongoing research on human pose retrieval in natural images. In~\cite{ferrari2009pose}, human upper-body pose estimation is combined with pose retrieval in a single pipeline. The estimated pose distributions are processed (\eg marginalized over locations) to obtain descriptors which can be directly compared through the defined distance. When multiple poses are used as a query, distance to the hyperplane estimated by SVM is used. The same approach to pose retrieval is used in~\cite{eichner20122d}, but with a modified pose detection. Both~\cite{ferrari2009pose,eichner20122d} consider only near-vertical upper-body poses with a focus on the frontal and rear view, as the main use-case is pose estimation and retrieval in frames from TV shows. These methods do not consider the full body pose and as such were confirmed to be inferior in preliminary experiments.

An alternative approach to pose retrieval is taken by~\cite{ren2012visual} where poses are represented using visual words without explicit pose estimation. First, bounding boxes of humans are detected. These are divided into a spatial grid in multiple scale levels. Each window is described by a vector quantized into visual words, so that each detected human has associated a sequence of visual words, encoding both its appearance and spatial layout. The task of matching poses across low-resolution archive of dance footage is then achieved by measuring the similarity between corresponding visual sentences. This method, however, encodes appearance rather than the pose alone and as such, it is not applicable in our scenario.

All described methods for pose retrieval, given a pose, retrieve images with a figure in a similar pose. In composition transfer discovery, the query is an image that can contain multiple figures. In that case, the task is to retrieve a similar spatial configuration of similar poses.

\subsection{Human pose estimation}

For the purpose of pose retrieval in this work, a trained pose detector is utilized to provide poses of humans in an image. There are many detectors with published code available; the most popular include OpenPose~\cite{cao2018openpose} and AlphaPose~\cite{fang2017rmpe}. Another option is to utilize semantic segmentations instead of pose detections and for this purpose, Mask R-CNN~\cite{he2017maskrcnn} was tested. During our preliminary experiments, OpenPose performed the best on our task and was chosen to provide pose detections on paintings.

\begin{figure}
    \centering
    \includegraphics[width=1.945in]{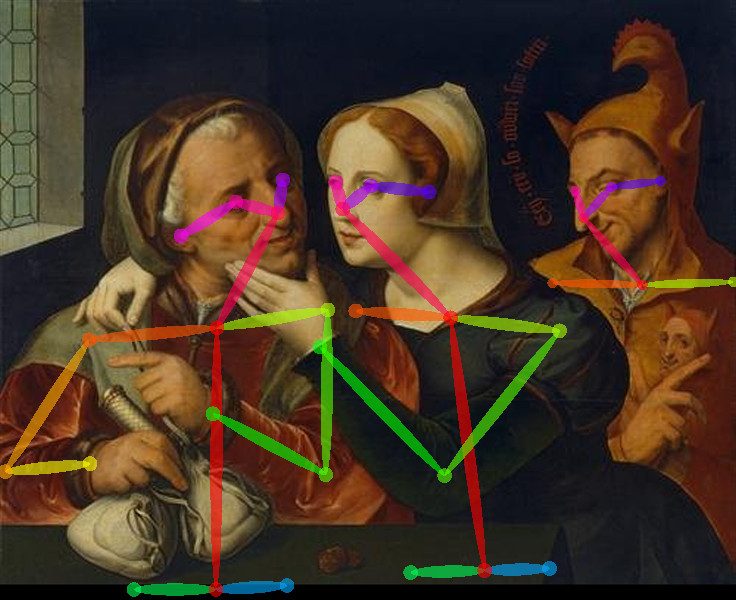} \hspace{0.2cm}
    \includegraphics[width=1.2in]{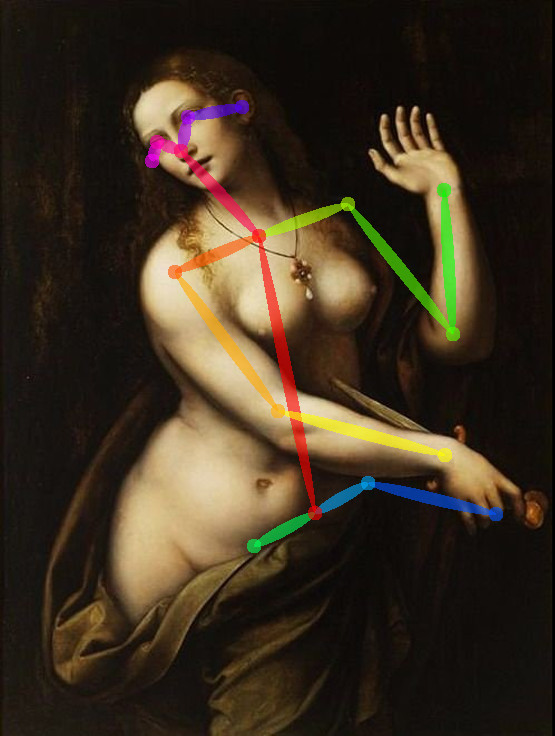}
    \caption{A visualization of 2D poses output by the OpenPose detector~\cite{cao2018openpose}. Each connection of a pair of detected keypoints is visualized in a different color. Missing and incorrect keypoint detections can be visible in the left and right image respectively.}
    \label{fig:detections}
\end{figure}

OpenPose~\cite{cao2018openpose} uses a multi-stage CNN architecture followed by a part association step to output 2D pose estimations for humans in the image. The output of the CNN is iteratively refined in each stage. For the first half of the architecture, the output is a part affinity field which is a 2D vector field encoding the degree of association between parts of the image. The output of the second half is a set of confidence maps, one for each body part. Each confidence map defines locations of body part detection candidates. The approximation of the most probable association of body part candidates according to the part affinity field is then found with a greedy approach. This provides an estimation of pose keypoint locations and associated confidence for each human in the image. Keypoints together with connections between pairs of them can be used to visualize a pose skeleton as in Figure~\ref{fig:detections}.

\section{Pose Retrieval and Verification}

The proposed pose-based retrieval system of planar visual art consists of two steps -- pose matching and geometric verification. The system works with pose detections, i.e. estimated poses of all humans in database images. For a query image, a fast and scalable method is used to identify potential matches. Poses in the images are compared independently, so the relations between individual poses are not considered in this step. Based on a rough geometric alignment of individual figures, the fast matching step generates a shortlist of images that are potentially similar to the query image, together with tentative individual pose matches.

As a second step, a robust spatial verification is performed. For each figure-to-figure hypothesis from the initial ranking phase, a geometric transformation between the query and the result image is estimated. The transformation is used to align the two images, \ie align simultaneously all the figures in them, and the number of pose keypoints consistent with it is verified. This step is robust to missing or misplaced keypoints and allows to measure similarity among more difficult scenes with multiple people.

When a human figure is re-used in a different artwork, it is either scaled or its position is shifted, but is never tilted or scaled in the two axis differently. 
Therefore, in both steps, the geometric transformation has three degrees of freedom: isotropic scale and translation.
When an engraving was created for a painting, the resulting print is flipped around the vertical axis and so are the human figures in every painting that was created from that print. The horizontal flip invariance of the system is explicitly considered.

\subsection{Fast matching}

Before fast pose matching, all poses are normalized. This ensures that the matching is shift-invariant to the pose position. The normalization is performed by choosing a root keypoint and making it the new origin, i.e. subtracting its coordinates from all other keypoints. The point directly between shoulders, labeled neck, was experimentally shown to be the most robust with respect to missing and noisy detections and was chosen as the root keypoint.

The pose detector outputs a $k$-tuple $r \in \mathbb{R}^{2 \times k}$ of coordinates $r_i$, $r = (r_1,\ldots,r_k)$ of specific pose keypoint locations ($k = 25$ for the detector used). 
During pose detection, not all keypoints of a pose may be detected. This happens when a part of the body is not recognized, occluded, or out of picture. In order to make the pose distance robust to missing keypoints, it considers detected keypoints only. Whether kyepoints are detected is indicated by the detector.

For two poses $r, s \in \mathbb{R}^{2 \times k}$, the pose distance is computed based on keypoints detected in both poses.
Let the set $I_{r,s}$ identify indices of these keypoints
\begin{equation}
I_{r,s} = \{i \mid r_i \text{\;is\;detected} \wedge s_i \text{\;is\;detected}, i \in \{1, .., k\} \} \mbox{.}
\label{eq:I}
\end{equation}
Translation invariant representation $r'$ and $s'$ respectively is obtained by subtracting the root
\begin{equation}
r'_i = \begin{cases}r_i - r_\text{root} & \text{if } i \in I_{r,s} \\ 0 & \text{else}\end{cases} \mbox{.}
\label{eq:r}
\end{equation}
The pose distance is computed as the inverse cosine similarity between the normalized pose vectors
\begin{equation}
\text{p}(r, s) = 1 - \frac{\langle r', s' \rangle}{\|r'\| \|s'\|} \mbox{,}
\label{eq:p}
\end{equation}
where $\langle r', s' \rangle$ denotes a dot product of vectorized representations, $\mathrm{vec}(r')^\T \mathrm{vec}(s')$. The normalization $\|r'\|$ can be geometricly interpreted as an estimate of scale, constraining the sum of squared distances of keypoints $I_{r,s}$ to 1; the dot product is proportional to the sum of squared distances of the scale-normalized keypoints.

The horizontal flip invariance is handled explicitly by defining the mirror-invariant pose distance $q$ where $s^*$ is pose $s$ mirrored around the vertical axis.

\begin{equation}
\text{q}(r, s) = \min \{p(r, s), p(r, s^*)\}
\end{equation}

To compute similarity between two images $i_1$ and $i_2$, we propose and evaluate two approaches that combine the distances of individual poses. In the baseline approach, the image similarity is given by the best matching pose as the minimum distance over all pairs of poses 
\begin{equation}
\text{dist}_\text{min}(i_1, i_2) = \min_{r \in i_1, s \in i_2} \text{q} (r,s)\mbox{.}
\label{eq:dist_min}
\end{equation}
As an alternative approach, an approximation of the maximum bipartite matching of poses in the two images is computed as
\begin{equation}
\text{dist}_\text{t}(i_1, i_2) = \sum_{r \in i_1} \min \{t,\ \min_{s \in i_2} \text{q} (r,s)\}\mbox{,}
\label{eq:dist_t}
\end{equation}
where $t \in \mathbb{R}$ is a maximum penalty for a non-matching pose.

Given the distance of each database image to the query image, $l$ images with the smallest distance are kept in the shortlist.

\subsection{Robust verification}

After potential matches are identified, geometric validation is performed to filter out images for which poses cannot be aligned with the query image. The pose bipartite matching between a pair of images is unknown, so all pose pairs matched during the previous step are considered.

For each tentative pose correspondence (one figure in the query image, one figure in the database image), a geometric transformation is estimated. The transformation consists of scale, translation and horizontal flip. Using two keypoint correspondences, the transformation is estimated in terms of least-squares\footnote{An exact solution to the system of equations does not exist as the transformation has three degrees of freedom and there are four equations resulting in an overdetermined system.}. To achieve a horizontal flip invariance, an additional transformation is estimated using horizontally flipped query image pose and the transformation with a smaller error on the two keypoint correspondences is chosen.

The transformation is found with RANSAC~\cite{fischler1981random}, so that it has the largest number of inliers, i.e. keypoints consistent with the transformation. A pair of keypoints is considered consistent with a transformation when the keypoint from the potential image match, after projection to the query image, is within a specified distance from the query image keypoint. This threshold distance is relative with respect to the estimated query image pose size and is therefore different for each pose in the query image. Once a transformation with a sufficient number of inliers is found, all keypoint correspondences consistent with it are used to re-estimate the transformation in terms of least squares.

The output of the RANSAC method is the best transformation found, measured by the number of inliers. If the number of inliers is sufficient, the corresponding pose pair is considered validated, otherwise, the transformation is filtered out. Each transformation, corresponding to a different pair of poses, is applied on all validated pairs of poses, transforming other figures in the image. The maximum number of keypoints consistent with a transformation is used as a measure of image similarity.

\begin{figure*}
    \centering
    \includegraphics[width=1.22in]{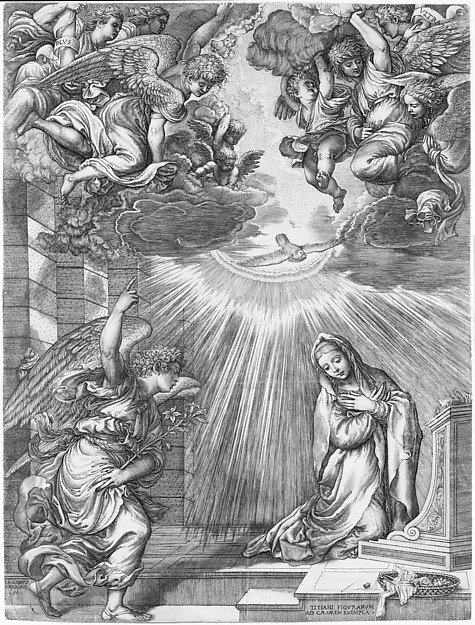}
    \begin{tikzpicture}
    \draw[white] (0,0) rectangle (0.15,3);
    \draw[->,thick] (0,2) -- (0.15,2);
    \end{tikzpicture}
    \includegraphics[width=1.95in]{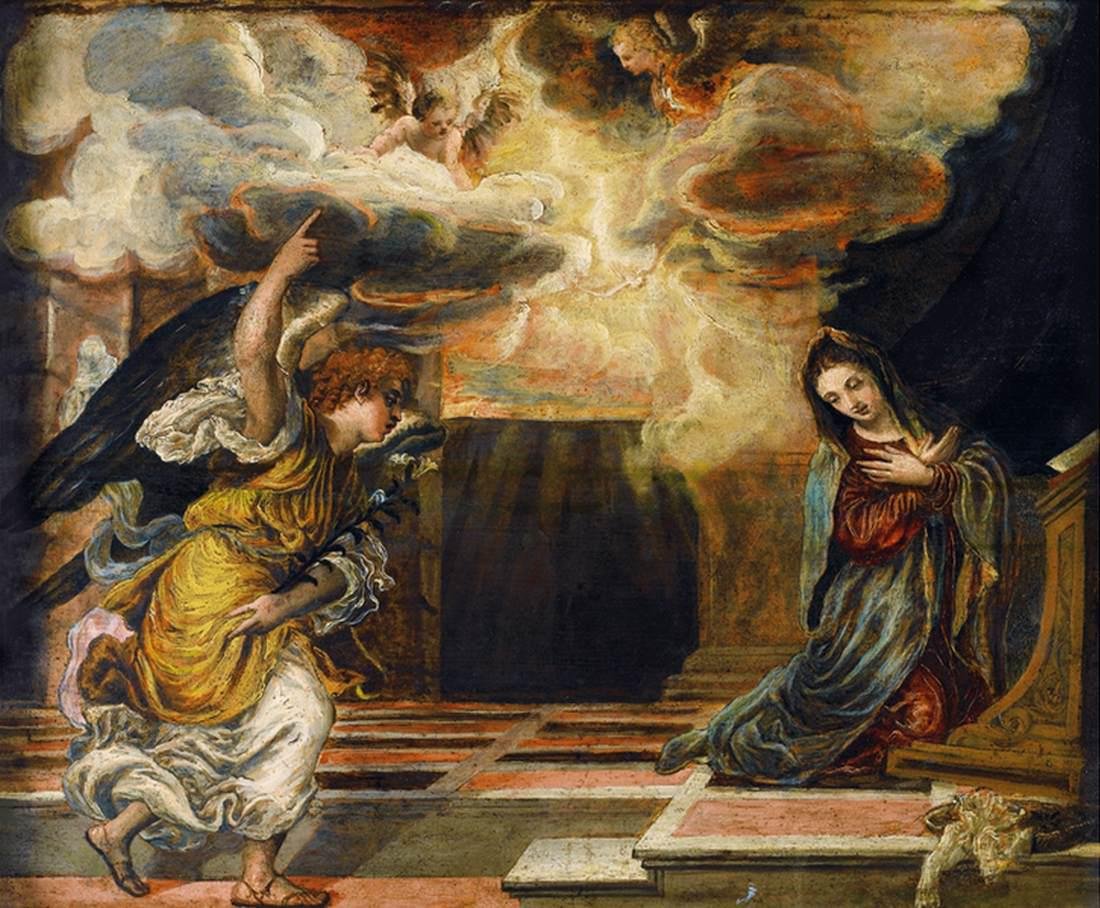} \hspace{0.2in}
    \includegraphics[width=1.35in]{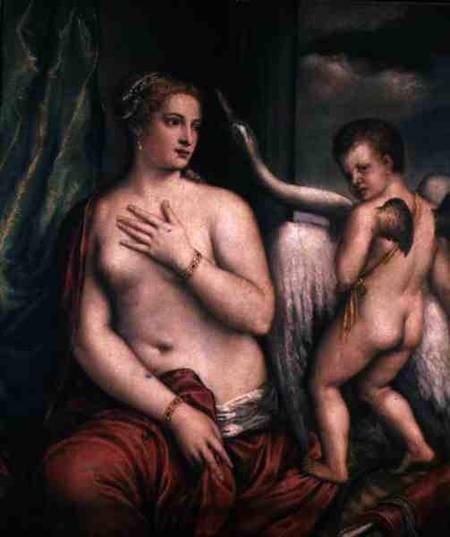}
    \begin{tikzpicture}
    \draw[white] (0,0) rectangle (0.15,3);
    \draw[->,thick] (0,2) -- (0.15,2);
    \end{tikzpicture}
    \includegraphics[width=1.35in]{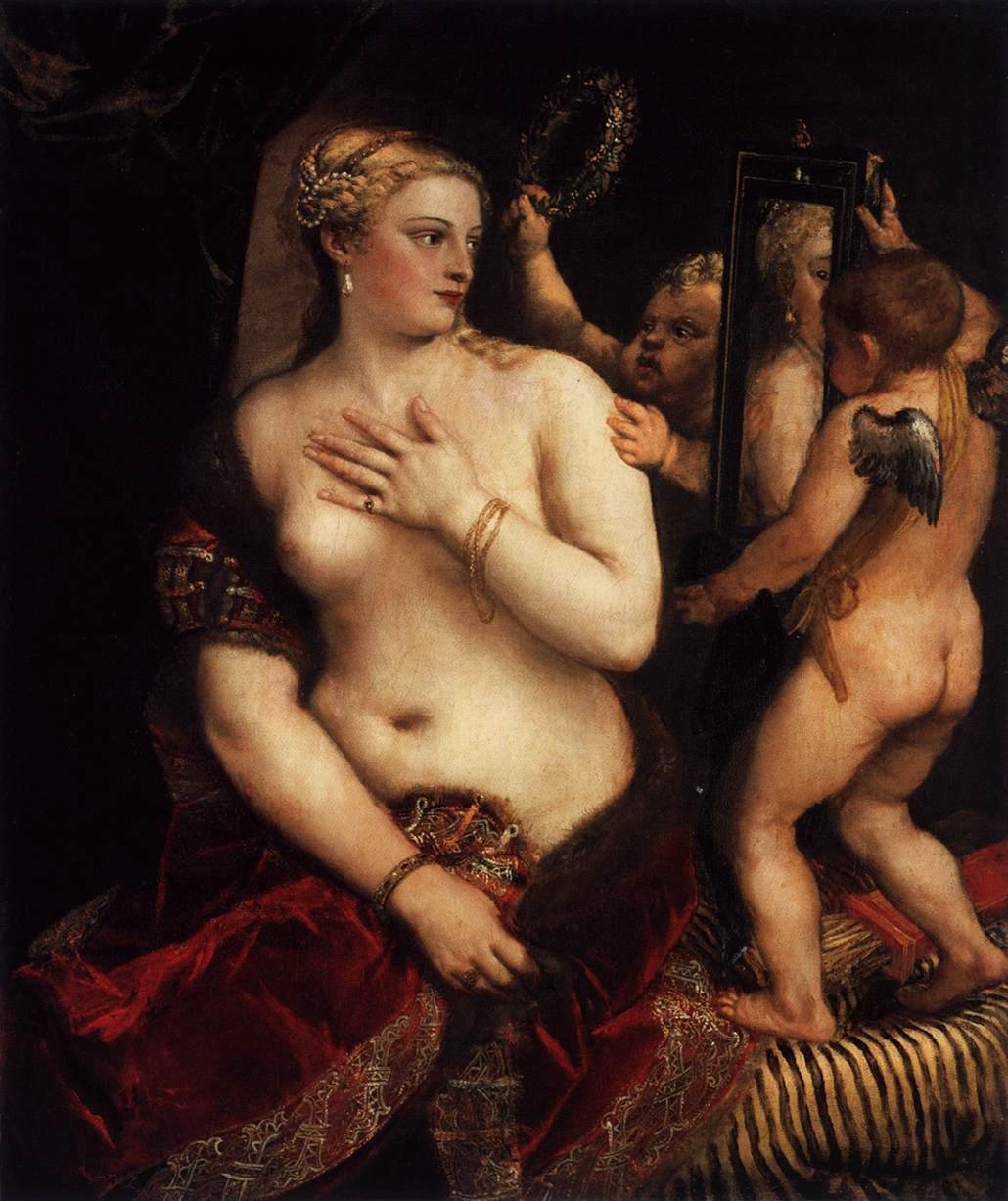}

    \vspace{0.25in}

    \includegraphics[width=1.69in]{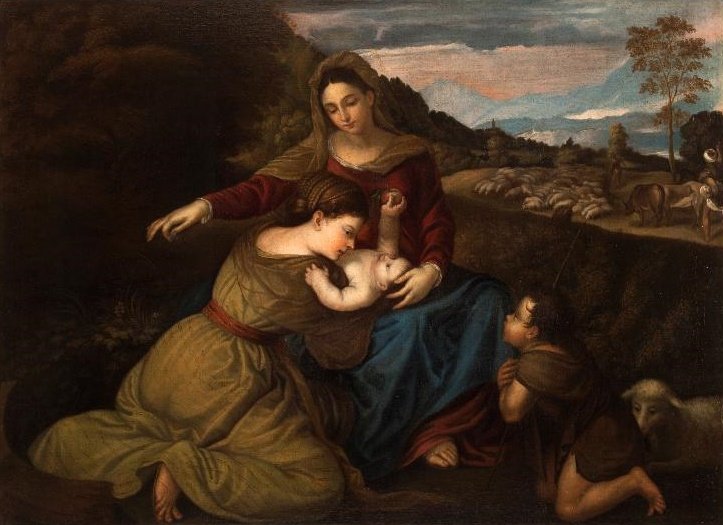}
    \begin{tikzpicture}
    \draw[white] (0,0) rectangle (0.15,3);
    \draw[->,thick] (0,1.5) -- (0.15,1.5);
    \end{tikzpicture}
    \includegraphics[width=1.75in]{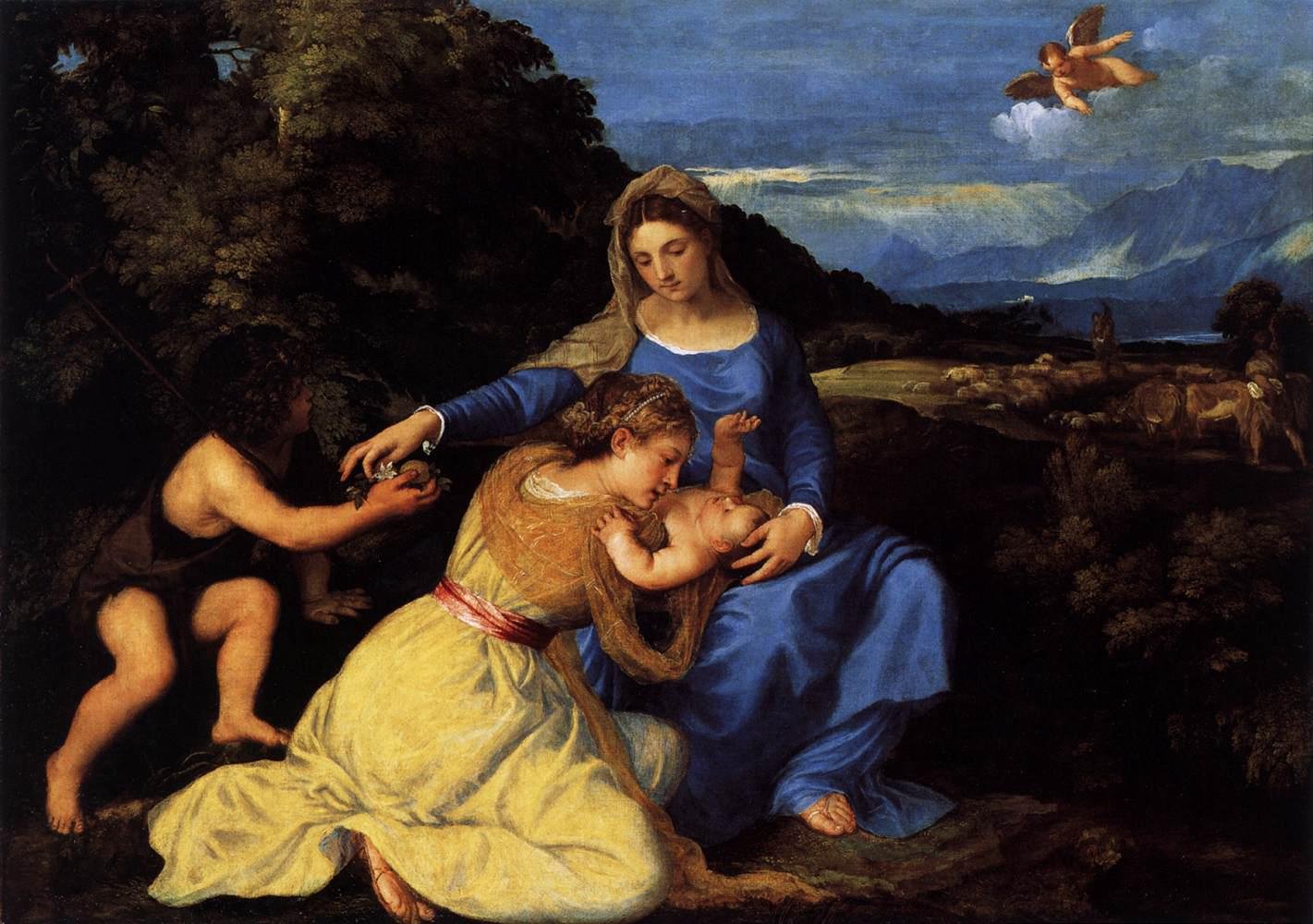} \hspace{0.25in}
    \includegraphics[width=0.64in]{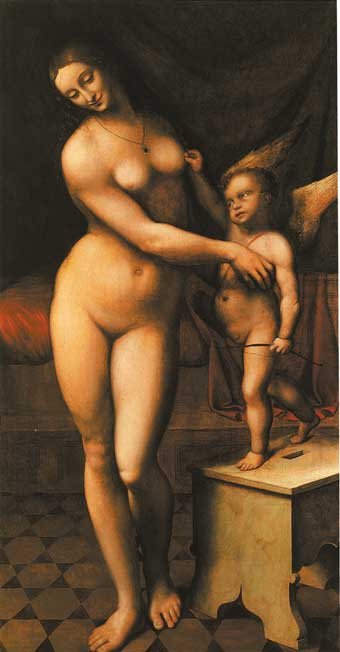}
    \begin{tikzpicture}
    \draw[white] (0,0) rectangle (0.15,3);
    \draw[->,thick] (0,1.5) -- (0.15,1.5);
    \end{tikzpicture}
    \includegraphics[width=0.75in]{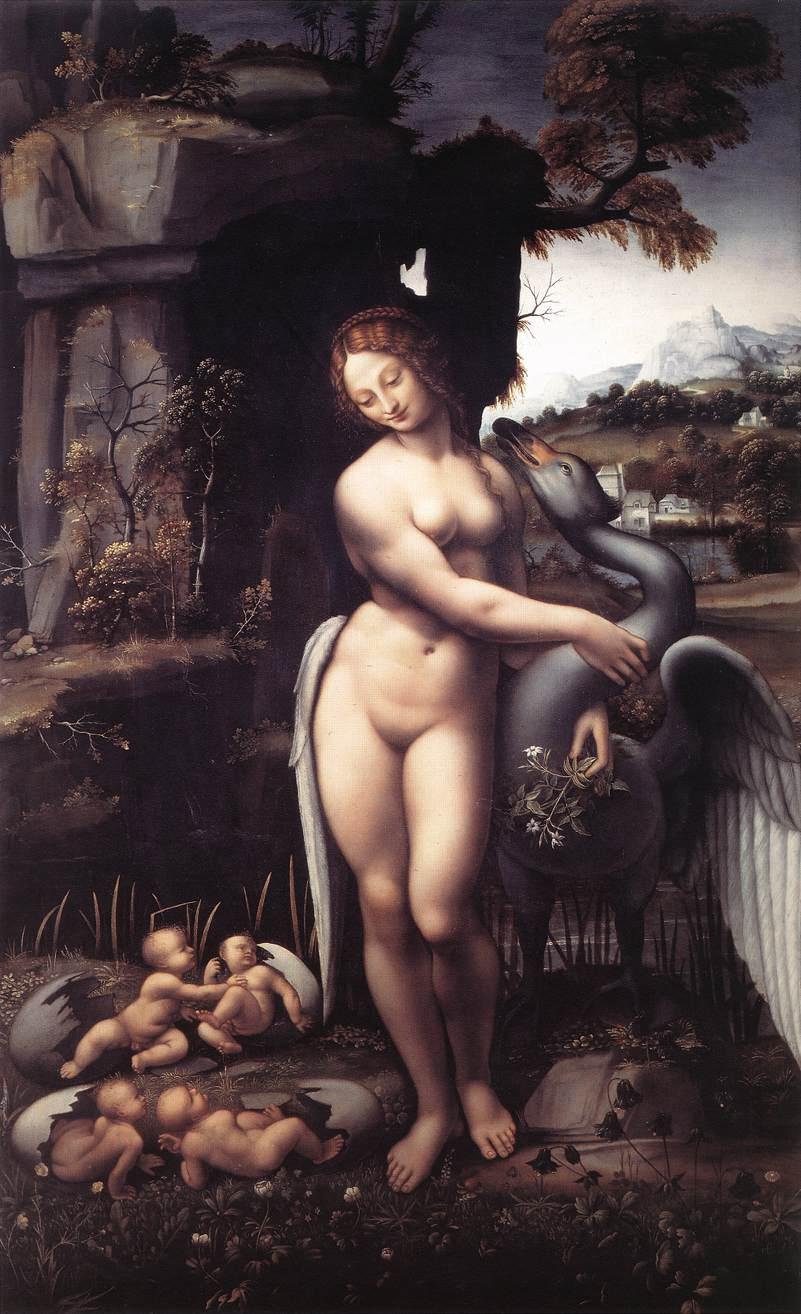}
    \includegraphics[width=0.91in]{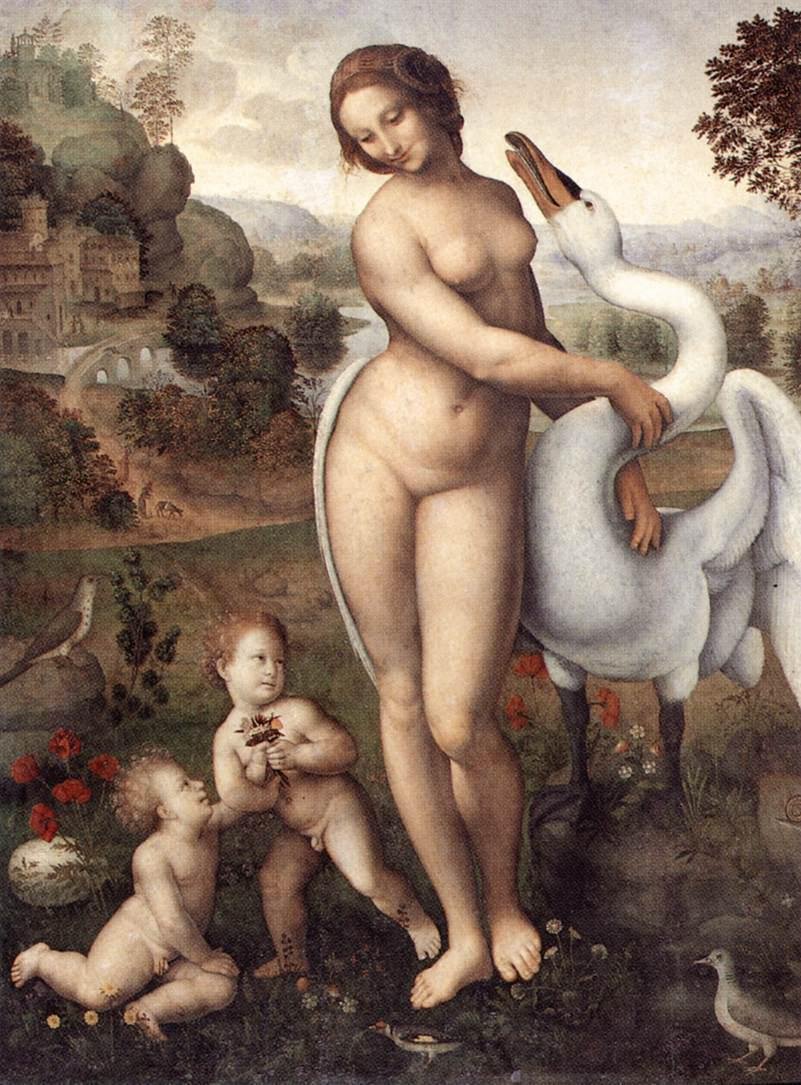}

    \vspace{0.25in}

    \includegraphics[width=1.77in]{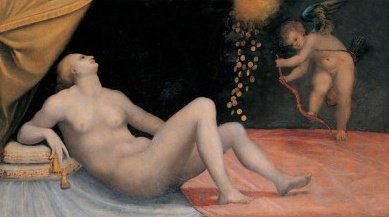}
    \begin{tikzpicture}
    \draw[white] (0,0) rectangle (0.15,2);
    \draw[->,thick] (0,1.2) -- (0.15,1.2);
    \end{tikzpicture}
    \includegraphics[width=1.4in]{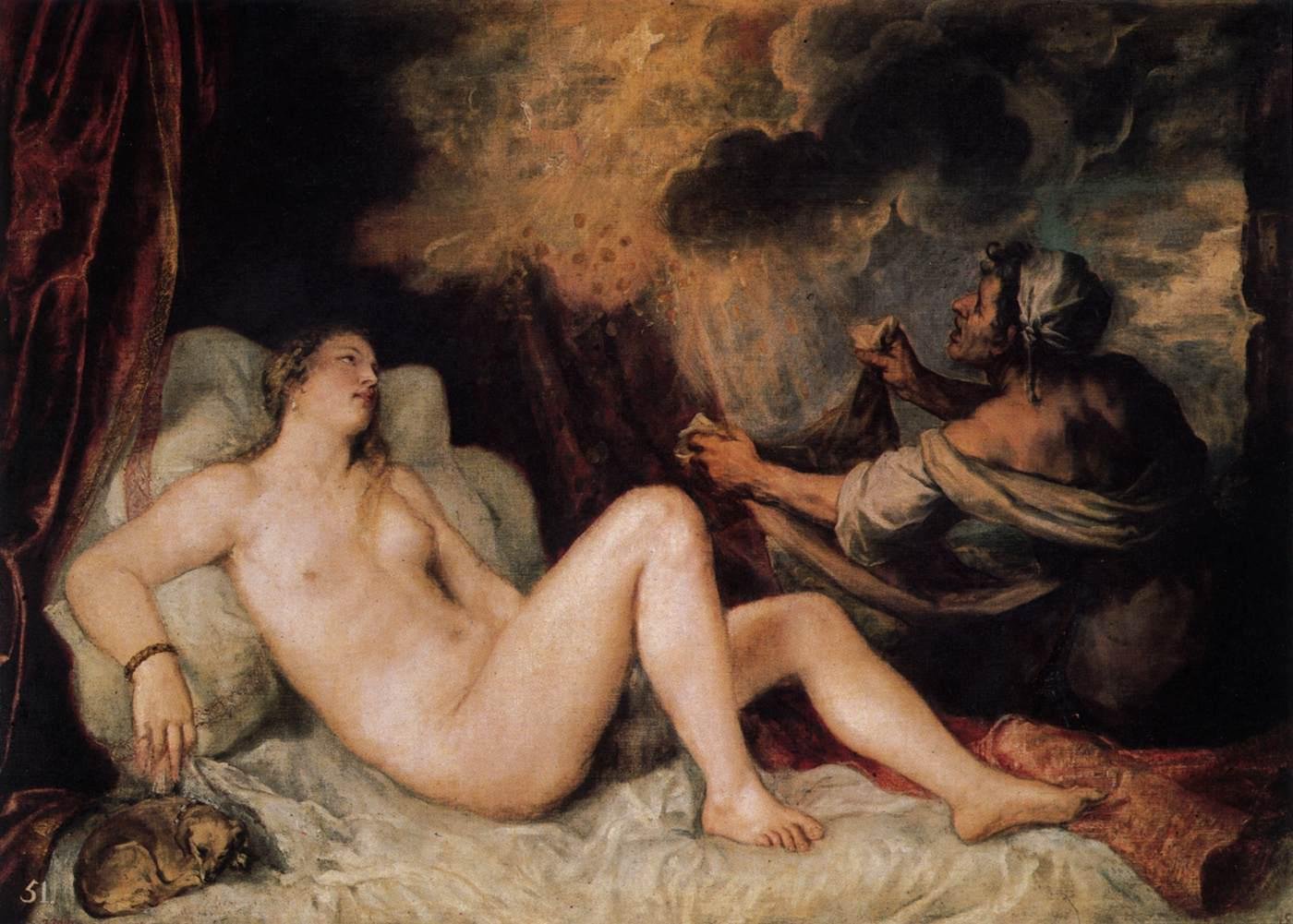}
    \includegraphics[width=1.46in]{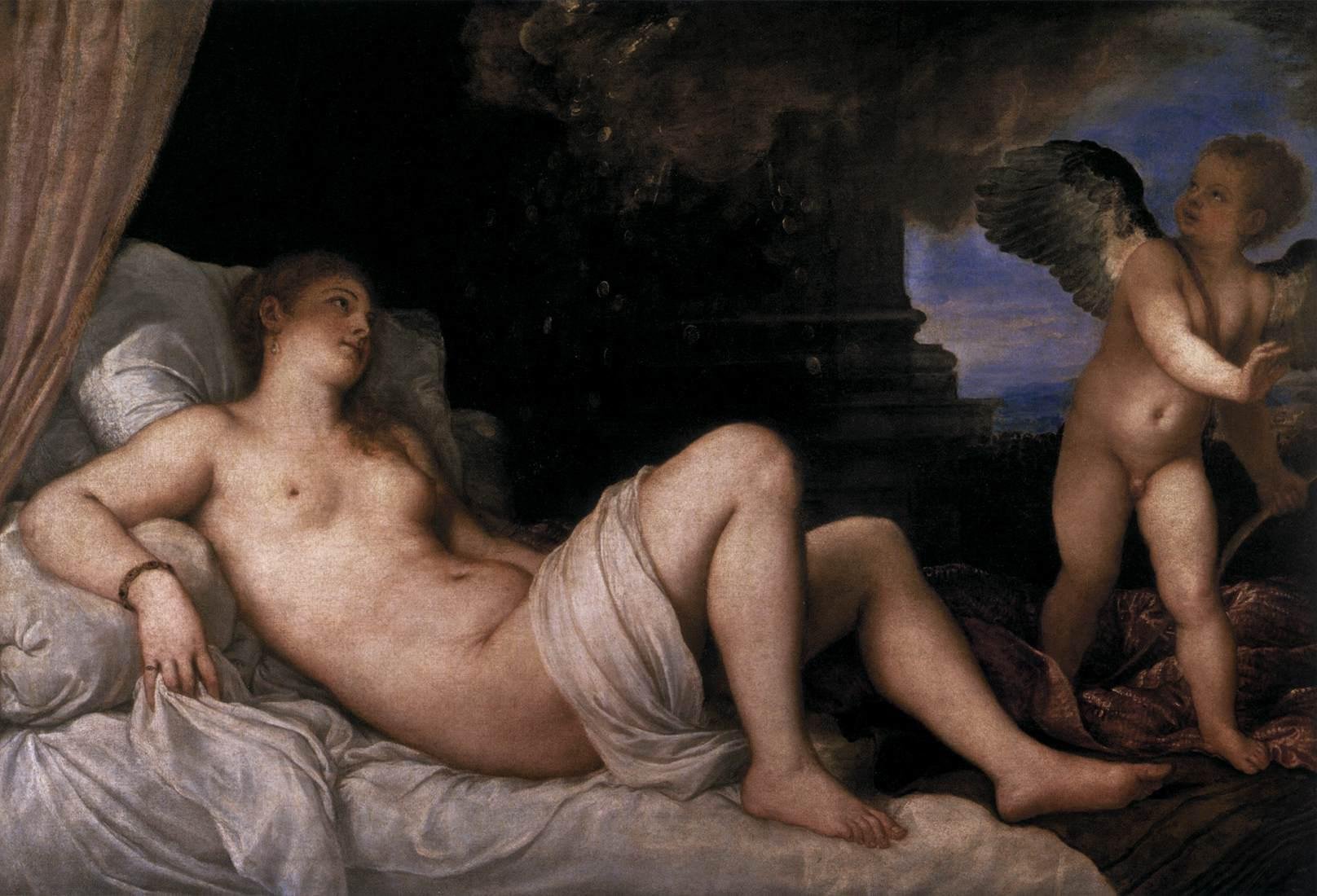}
    \includegraphics[width=1.57in]{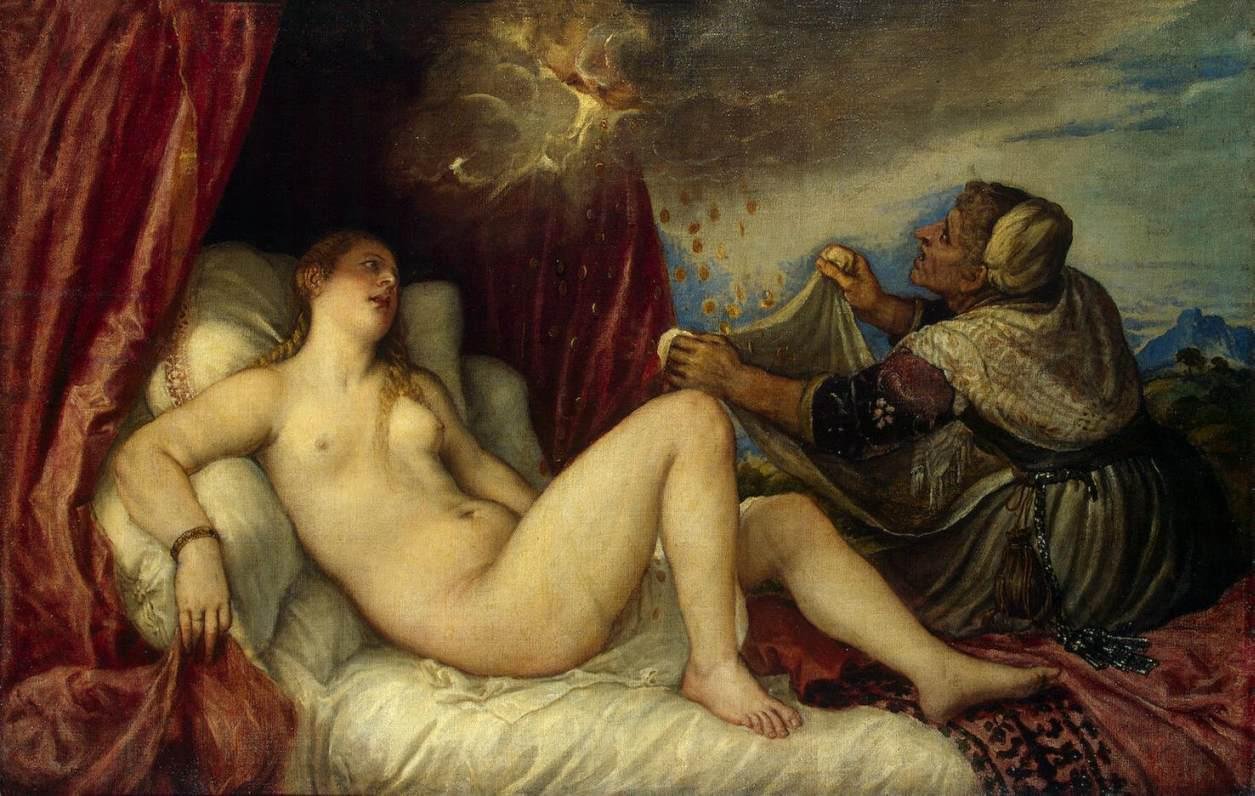}
    \caption{Examples of discovered links between artworks in the WGA database that are impossible to retrieve using standard image retrieval methods because of a substantial change in their visual appearance. The first image, corresponding to the query, is from the composition transfer dataset, the remaining images are from WGA in the order they were retrieved by our method.}
    \label{fig:discoveries}
\end{figure*}
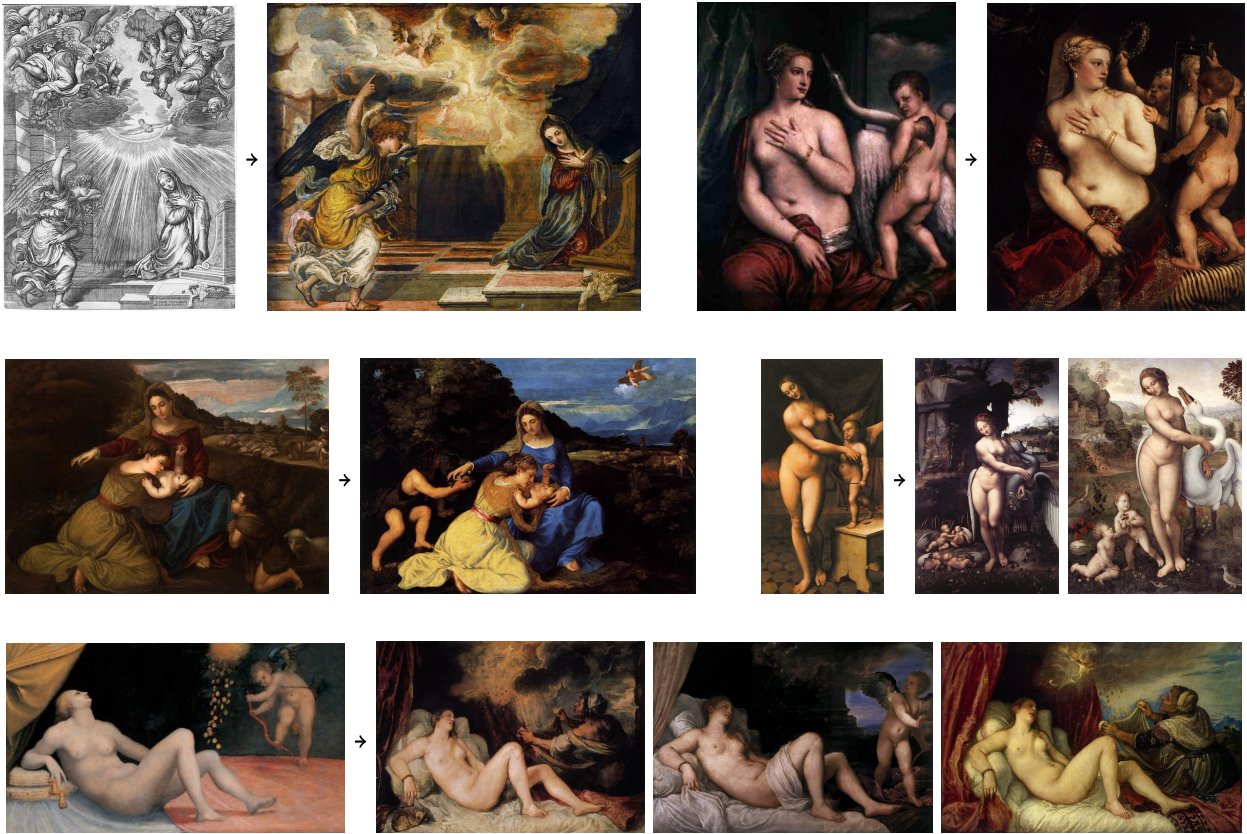

To determine the inlier threshold for RANSAC, relative query pose size with respect to the canonical pose size is estimated.
For a canonical pose, the distances between connected pose keypoints are known. The relative query pose size is computed as a median of the ratios between distances of connected keypoints detected in the query pose and corresponding distances in the canonical pose.

For efficiency reasons, a torso (the connection of neck and mid-hip) angle test is performed, and poses with the mutual torso angle larger than a threshold are eliminated as non-matching. This test is applied only to poses with both keypoints of the torso detected.

\section{Experiments}

In the experiments, we evaluate the proposed method both qualitatively and quantitatively. We compare against two CNN-based baseline approaches, namely VGG GeM~\cite{Radenovic-TPAMI18} and EdgeMAC~\cite{Radenovic-ECCV18}, which both embed images into a descriptor space. The first one, VGG GeM, is a state-of-the-art specific object image retrieval CNN that uses the RGB image to compute the descriptor. 
The latter one, EdgeMAC, is a method for image retrieval that extracts the descriptors from edge maps, and was shown invariant, to some degree, to the visual domain. For the proposed approach, we analyze the impact of two different image distance measures, and the importance of geometric validation. To evaluate the performance of the composition transfer discovery, we created a manually annotated art composition transfer dataset.

\subsection{Experimental setup}

Poses are detected with a state-of-the-art pose detector -- we use the OpenPose~\cite{cao2018openpose} body\_25 model from the official project site\footnote{https://github.com/CMU-Perceptual-Computing-Lab/openpose}. Other models provided by the authors, i.e. coco and mpi, were also tested but they were inferior to the body\_25 model. We follow the configuration recommended by the authors: 
each image is downscaled to have the longer edge equal to 736px and a pyramid of four scales, each being four times smaller then the previous one, was used as the input to the pose detector. 

In all experiments, we use a shortlist of $l = 50$ images as the input for geometric validation. This is a compromise between precision and speed of the validation whose time complexity is linear in the length of the shortlist. In this scenario, a non-optimized Python implementation of geometric validation took on average $4.5$~s for each query. The pose matching, whose time complexity is linear in the database size, took $100$~ms for each query for a database of 635 images with this implementation.

In the experiments, the following parameter setting is used, nevertheless, the method is not sensitive to a modest adjustment of these parameters.
For the evaluation of image distance dist\textsubscript{t}, we set $t = 0.05$. In geometric validation, pose pairs whose distance exceeds $0.1$ and whose torso angle exceeds $0.4$~rad are discarded.
The poses are considered validated if the estimated transformation aligned at least 1/4 of all keypoints of the pose, which is 7 out of 25 in our setup.
The parameter and threshold values were hand-tuned on a separate small validation set, but a more comprehensive parameter search would certainly increase the performance if larger dataset would be available.

\subsection{Datasets}

In order to evaluate the proposed approach and allow further comparison, we created a manually annotated dataset of 635 images. These images were hand-picked and downloaded from an unknown internet source by~\cite{seguin2018making}. The dataset consists of a set of paintings with repeating motifs and for some paintings, also associated drawings and engravings. It was assembled by people with an art history background as a set of artworks, mainly from renaissance and baroque, that is known to contain composition transfer. It includes, for example, artworks that are known to be frequently copied. To the best of our knowledge, there is no other publicly available dataset for our task.

The set of downloaded images is annotated by assigning each link between a pair of images one of two connection labels -- \emph{copy} or  \emph{composition transfer}. Physical links and replications were both labeled as \emph{copy} because, for some instances, distinguishing between the two based just on image data is ambiguous as the visual information is nearly identical in both cases. When this is the case, the pair of images offer the same challenge for image retrieval systems regardless their ground truth connection. The \emph{copy} category contains all images where the author tried to copy the painting and all media of the same painting, therefore covering a wide range of visual diversity between images. In the \emph{composition transfer} category, all visually different images with obvious theme inspiration were placed. Theme inspiration could be either in the pose of one of the characters or the configuration of multiple characters.

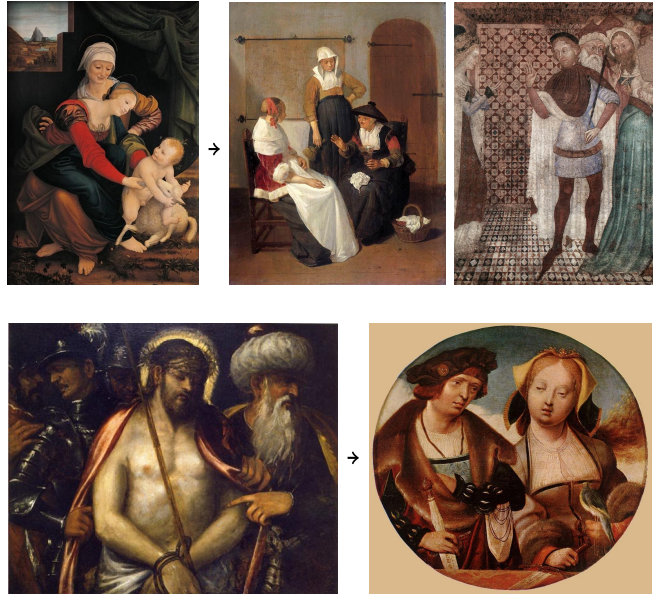
\begin{figure}
    \centering
    \includegraphics[width=1in]{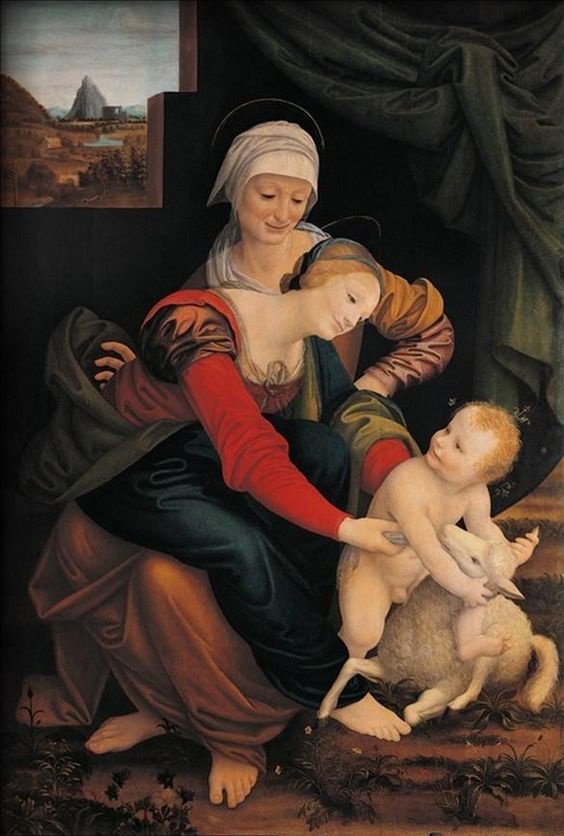}
    \begin{tikzpicture}
    \draw[white] (0,0) rectangle (0.15,3);
    \draw[->,thick] (0,1.8) -- (0.15,1.8);
    \end{tikzpicture}
    \includegraphics[width=1.13in]{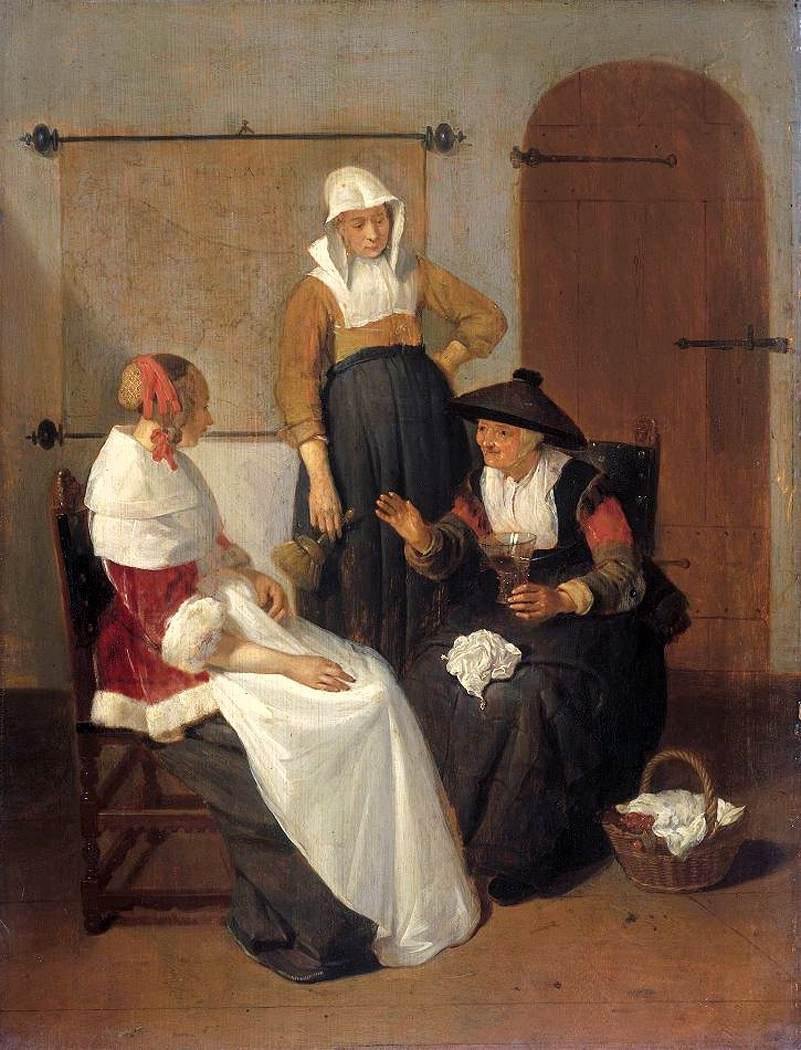}
    \includegraphics[width=1.04in]{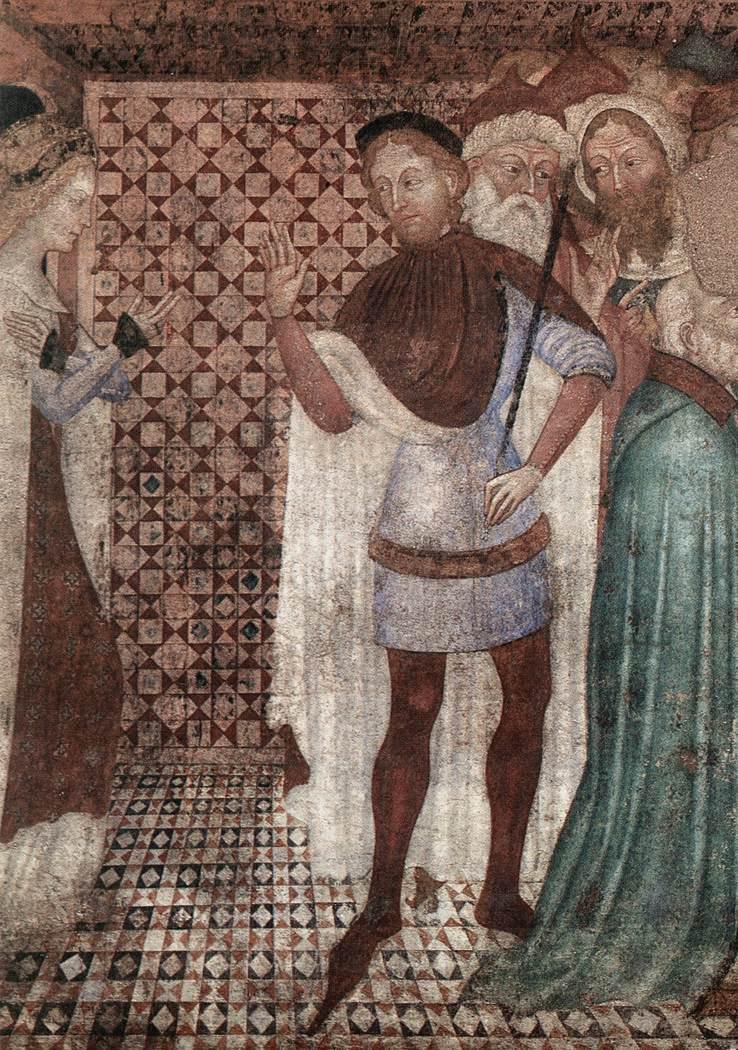}

    \vspace{0.2in}
    
    \includegraphics[width=1.72in]{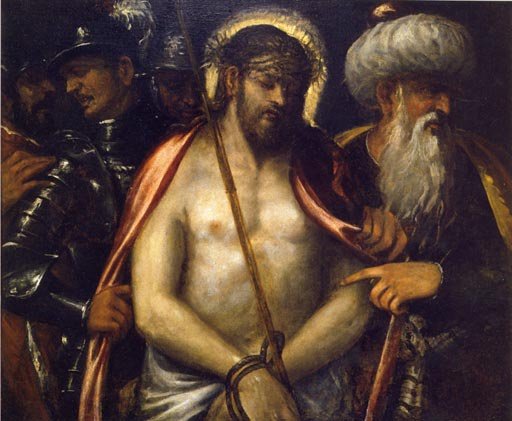}
    \begin{tikzpicture}
    \draw[white] (0,0) rectangle (0.15,3);
    \draw[->,thick] (0,1.8) -- (0.15,1.8);
    \end{tikzpicture}
    \includegraphics[width=1.48in]{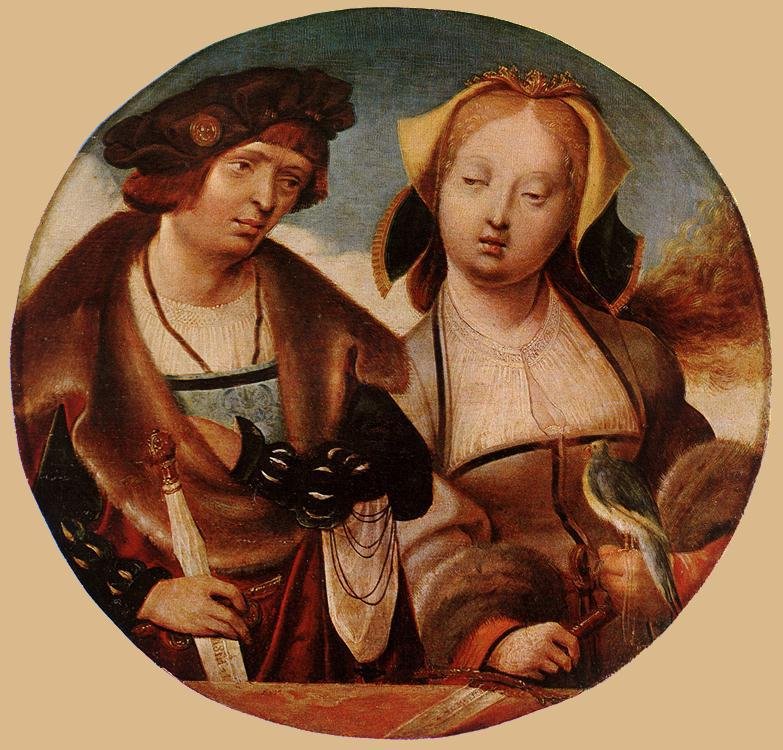}
    \caption{An example of false positives retrieved by our method: incorrect validation through a distinctive pose of one of the characters (top row) and through a generic pose of multiple characters amplified by noisy pose keypoint detections (bottom row). In all cases a query image from the composition transfer dataset (left image) was used to query the WGA database (remaining images) and visualized retrieved images were validated by our method with high confidence.}
    \label{fig:failures}
\end{figure}

The composition transfer dataset provides a simplified benchmark for theme adoption discovery among artworks. Nevertheless, the annotation is an expensive process requiring an art history insight as it involves painting categorization for the purpose of connection discovery. Therefore, only a qualitative evaluation on a bigger dataset is performed. We use the composition transfer dataset for query images and we search in 37 thousand images from the digitized collection of WGA which are publicly available\footnote{https://www.wga.hu/database/download}. We then visually inspect the results for each query and provide some interesting links discovered by our method in Figure~\ref{fig:discoveries}.

\subsection{Results}

We provide quantitative results using the composition transfer dataset for two evaluation scenarios. The performance is measured by mean precision at $i$-th rank (mP@$i$) for ranks ranging from 1 to 10. 
Results for two baselines and two versions of the image distance measure in the proposed method are reported.

In the first scenario, both labels -- \emph{copy} and \emph{composition transfer} -- are treated as positive. The results are summarized in Table~\ref{tab:results}. We also report the performance of our method with no geometric validation, to highlight the importance of that step.
In order to analyze the performance, in the second scenario, we separate the connection types, one is taken as positive and the other one is ignored in the evaluation. In Table~\ref{tab:copy_annotation}, we present the performance per connection type. 

The proposed method with image distance dist\textsubscript{t} provides the best results among all tested approaches on the composition transfer dataset. It outperforms both VGG GeM and EdgeMAC image retrieval baselines by a significant margin. When the validation is skipped and only matching is performed in our method, it is inferior to both baselines, with dist\textsubscript{t} providing significantly better results than dist\textsubscript{min}.

In general, all methods provide better results on the \emph{copy} connections than on the \emph{composition transfer} connections. From the two baselines, EdgeMAC offer similar performance as VGG GeM when both connections are considered, but it can be seen that their results differ with respect to the connection type. EdgeMAC is superior to VGG GeM on \emph{copy} connections but inferior in \emph{composition transfer} discovery. This is caused by the fact that in \emph{copy} connections, the same artwork across different media often occurs, and their visual difference is decreased by converting the images to edge maps.

In our method, the approximation of the maximum bipartite matching of poses implemented by the image distance dist\textsubscript{t} performed only slightly better than simply taking a pose distance minimum with dist\textsubscript{min}. This appears to be inconsistent with the results when performing matching only. It can be seen in Table~\ref{tab:results} that dist\textsubscript{t} significantly outperforms dist\textsubscript{min} when performing matching only, most notably for mean precision at rank 1 (mP@1).
When geometric validation is performed, the poses are explicitly matched and the performance gain of matching the poses in dist\textsubscript{t} in the preceding step is substantially decreased. There still remains a subtle performance gain of dist\textsubscript{t} as some images that could be geometrically validated do not get connected by dist\textsubscript{min} through individual poses and therefore do not get to the geometrical validation phase. This is most pronounced in images with multiple characters where the combination of characters' poses is more distinctive than each character's pose alone.

\begin{table}[t]
\centering
\begin{tabular}{|l||r|r|r|r|r|}\hline
Method & mP@1 & mP@2 & mP@5 & mP@10 & mP@50 \\ \hline \hline
VGG GeM~\cite{Radenovic-TPAMI18} & 56.8 & 51.0 & 41.7 & 34.6 & 26.9 \\\hline
EdgeMAC~\cite{Radenovic-ECCV18} & 59.3 & 52.4 & 42.3 & 35.0 & 27.3 \\\hline\hline
dist\textsubscript{min} {\scriptsize match-only} & 29.4 & 29.5 & 27.6 & 24.7 & 21.5 \\\hline
dist\textsubscript{t} {\scriptsize match-only} & 50.1 & 47.3 & 39.3 & 33.5 & 27.3 \\\hline
dist\textsubscript{min} & 74.6 & 68.3 & 57.3 & 49.8 & 40.1 \\\hline
dist\textsubscript{t} & {\bf 75.3} & {\bf 69.3} & {\bf 58.1} & {\bf 50.6} & {\bf 40.9} \\\hline
\end{tabular}
\caption{Performance comparison on all art connection annotations from created dataset, measured in temrs of mean precision}
\label{tab:results}
\end{table}

\begin{table}[t]
\centering

Copy Connections \vspace{0.03in}

\begin{tabular}{|l||r|r|r|r|}\hline
Method & mP@1 & mP@2 & mP@5 & mP@10 \\ \hline \hline
VGG GeM~\cite{Radenovic-TPAMI18} & 64.2 & 59.7 & 51.2 & 45.3 \\\hline
EdgeMAC~\cite{Radenovic-ECCV18} & 71.0 & 65.2 & 56.4 & 50.1 \\\hline\hline
dist\textsubscript{min} & {\bf 87.7} & 83.5 & {\bf 74.7} & 69.4 \\\hline
dist\textsubscript{t} & {\bf 87.7} & {\bf 84.0} & 74.6 & {\bf 69.6} \\\hline
\end{tabular}

\vspace{0.4cm}

Composition Transfer Connections \vspace{0.03in}

\begin{tabular}{|l||r|r|r|r|}\hline
Method & mP@1 & mP@2 & mP@5 & mP@10 \\ \hline \hline
VGG GeM~\cite{Radenovic-TPAMI18} & 41.0 & 36.0 & 30.4 & 25.5 \\\hline
EdgeMAC~\cite{Radenovic-ECCV18} & 40.4 & 34.1 & 28.2 & 24.3 \\\hline\hline
dist\textsubscript{min} & 55.7 & 50.4 & 41.2 & 35.6 \\\hline
dist\textsubscript{t} & {\bf 57.5} & {\bf 51.9} & {\bf 42.6} & {\bf 37.4} \\\hline
\end{tabular}
\caption{Performance comparison on copy (top) and composition transfer (bottom) art connection annotations from created dataset, measured in terms of mean precision}
\label{tab:copy_annotation}
\end{table}

\subsection{Failure cases}

\begin{figure}
    \centering
    \includegraphics[width=1.8in]{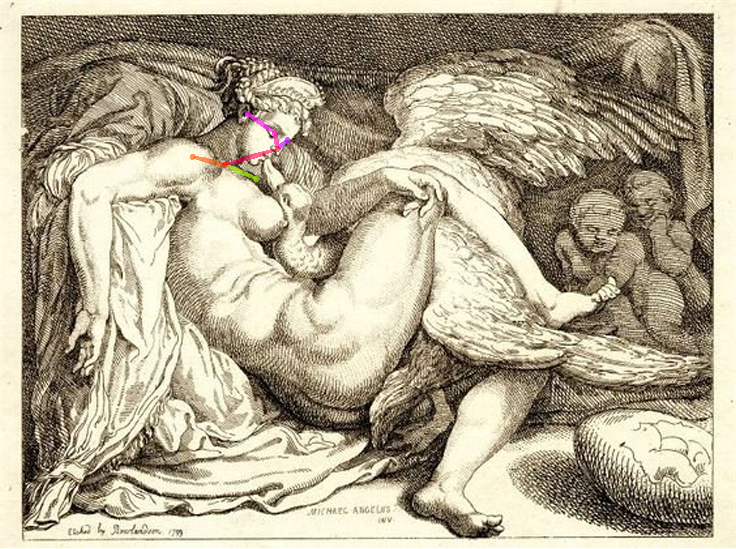} \hspace{0.1cm}
    \includegraphics[width=1.44in]{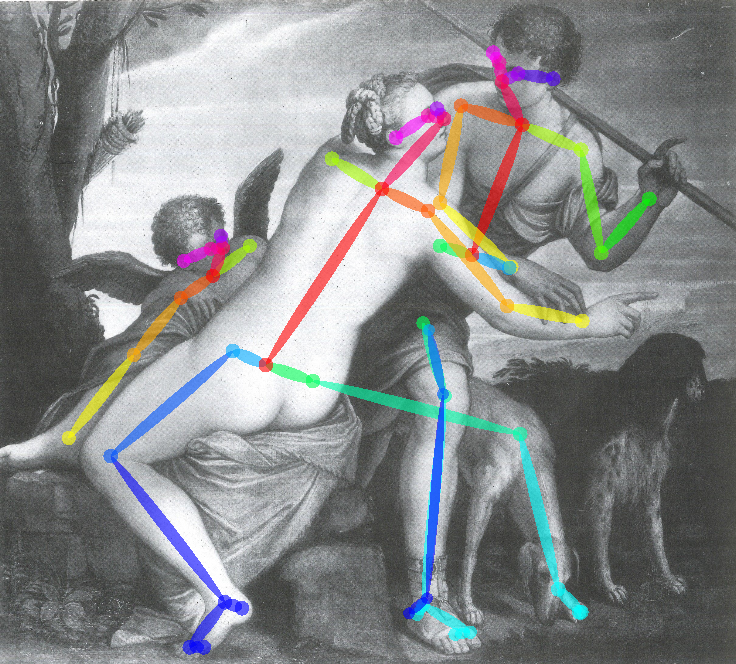}
    \caption{Failure cases of the OpenPose detector. In the left image, only the face is detected, whereas in the right image, correctly detected keypoints are associated with wrong poses. In both cases, the error is most likely caused by a combination of low-confidence detections with a prior on a human figure in the part affinity fields used for part association.}
    \label{fig:detection_failures}
\end{figure}

As our method relies on pose detections, failures in this stage prevent images to be correctly linked. The detection failures range from completely missing a pose, partial detections to construction of incorrect connections (\eg multiple people ocluding each other). Examples of pose detector failure cases are given in Figure~\ref{fig:detection_failures}.
We observed that our method is robust to missing keypoint detections as long as approximately two thirds of keypoints in a database image are detected. If a lower number of keypoints is detected, less relevant images start to match the query better. This is a minor issue for the composition transfer dataset, but for more recent modern-form art, such as cubism but also impressionism, the pose detection systematically fail. We were unable to quantify the accuracy of used pose detector on artworks as there is currently no artwork dataset with ground-truth pose annotations to the best of our knowledge.

Another issue is that the transformation does not consider a change in the mutual arrangement of characters. An example of this can be seen in the bottom row of Figure~\ref{fig:motivation} where the characters are moved closer to each other. In this specific case, both poses were matched individually and there was no better-matching image in the database, so all relevant images were retrieved. However, the geometrical validation step is not capable of validating the composition of the figures by finding a single unifying transformation. Another difficulty is pose ambiguity. Our method, in general, relies on characters having distinctive poses, which is common in the case of art. There is, nevertheless, an issue with paintings of crowds, which can be aligned with non-relevant paintings with high confidence. An example of incorrect validation through a distinctive pose of an individual figure and through a generic pose of multiple figures is provided in Figure~\ref{fig:failures}.

\section{Conclusions}

In this paper, a method linking paintings through similarity of depicted human poses and their arrangements in a large collection of digitized artworks was proposed. The method takes the output of a state-of-the-art pose detector and attempts to find similar poses for every image in the collection by a two-step retrieval procedure. First, candidate links are found by an efficient fast matching and then, precise geometric verification is performed on the candidate shortlist. It was experimentally shown, that the proposed method significantly outperforms current content-based image retrieval methods that attempt to measure general visual similarity.